\documentclass[12pt]{scrartcl}
\usepackage[T1]{fontenc}							
\usepackage[utf8]{inputenc}

\usepackage{geometry}
\usepackage{scrpage2}
	\automark[subsection]{subsection}
\usepackage{setspace}
	\spacing{1.2}
\usepackage{enumerate}
\usepackage{url}
\usepackage{array, multirow, dcolumn}
\usepackage{rotating}
\usepackage{booktabs}

\usepackage{xcolor, graphicx}

\usepackage{amssymb}
\usepackage{amsmath, amsthm}

\KOMAoption{headings}{small}

\DeclareMathOperator*{\argmin}{arg min}
\DeclareMathOperator*{\argmax}{arg max}

\providecommand{\keywords}[1]{\textbf{\textit{Keywords --- }} #1}

\theoremstyle{definition}

\begin{document}

\title{CITlab ARGUS for historical data tables}
\subtitle{Description of CITlab's System for the ANWRESH-2014 Word Recognition Task}
\author{Gundram Leifert \and Tobias Grüning \and Tobias Strauß \and Roger Labahn\thanks{corresponding author; CITlab, Institute of Mathematics, University of Rostock\newline\{gundram.leifert, tobias.strauss, tobias.gruening, roger.labahn\}@uni-rostock.de}}
\date{April 14, 2014 \\ last revision: May 02, 2014}
\maketitle

\begin{abstract} 
We describe CITlab's recognition system for the ANWRESH-2014 competition attached to the 14.\,International Conference on Frontiers in Handwriting Recognition, ICFHR 2014. The task comprises word recognition from segmented historical documents. The core components of our system are based on multi-dimensional recurrent neural networks (MDRNN) and connectionist temporal classification (CTC).
The software modules behind that as well as the basic utility technologies are essentially powered by PLANET’s ARGUS framework for intelligent text recognition and image processing.
\end{abstract}

\section{Introduction}
The International Conference on Frontiers in Handwriting Recognition ICFHR 2014 \linebreak (\url{www.icfhr2014.org}) hosts a variety of competitions in that area. Among others, \linebreak ANWRESH-2014 attracted our attention because we expected CITlab's handwriting recognition software to be able to successfully deal with the respective task.
With a system very similar to the one presented here, the CITlab team also took part in 2014 ICFHR's HTRtS competition on historical handwritten document recognition, see \cite{citlab2014htrts} for the according system description.

ANWRESH-2014 comprises a task of word recognition for segmented historical documents, see \url{http://collections.ancestry.com/ANWRESH-2014} for further details.
These data consist of page images taken from the 1930 US Census General Population Schedules.
Each page mainly has a table with 50 rows containing various fields with personal data, out of which five are to be read here.
Table \ref{tab:NetworkTypes} explains the essential data fields and shows the labels used throughout the paper to refer to the respective network type.
\begin{table}[h]
	\centering
	\begin{tabular}[t]{c@{\hspace{1em}}p{.7\linewidth}}
		network type \dots & reads data field \dots \\ \hline \\[-.75em]
		$\mathcal{N}$ & \texttt{NAME}, which itself is composed of family name and given name \\
		$\mathcal{R}$ & \texttt{RELATION}, e.g. \textsl{Head, Lodger, Wife, Daughter, \dots} \\
		$\mathcal{A}$ & \texttt{Age} (at last birthday), i.e. integers and possibly fractions counting month \\
		$\mathcal{M}$ & \texttt{Marital condition} \\
		$\mathcal{B}$ & \texttt{PLACE OF BIRTH}
	\end{tabular}
	\caption{Network types: Labels and Data}
	\label{tab:NetworkTypes}
\end{table}

Moreover, we apply committees of networks for mixture-of-experts evaluations for certain data. This variety of networks is referred to by indexing. 

Our neural networks have basically been used previously in the international handwriting competition OpenHaRT 2013 attached to the ICDAR 2013 conference, see \cite{citlab2013openhart}.
Moreover, with a system very similar to the one presented here, the CITlab team also took part in ICFHR's HTRtS competition on a handwritten text recognition task, see \cite{citlab2014htrts} for the according system description.

Affiliated with the Institute of Mathematics at the University of Rostock, CITlab\footnote{\url{http://www.citlab.uni-rostock.de}} hosts joint projects of the Mathematical Optimization Group and PLANET intelligent systems GmbH, a small/medium enterprise focusing on computational intelligence technology and applications.
The work presented here was part of a common text recognition project 2010 -- 2014 and is extensively based upon PLANET's ARGUS software modules.

\section{Preprocessing}
\label{s:Preprocessing}
For ANWRESH-2014, additional field segment information was provided but the given polygons typically do not match the table field borders exactly.
Therefore, we firstly tried to find the lines of the table more precisely: Starting from a certain enlargement of the given polygons as a rough region approximation, we select pixel density peaks in projection profiles to both page borders.  
Cutting-out along those table lines yields the single writings to be recognized.

Note that the neural networks only work properly as long as the input image properties are close to or even meet those of the images shown during training. In order to meet such requirements, and depending on the network type (see Table \ref{tab:NetworkTypes}), we first normalize the image heights to 96 or 128 pixels for networks of type $\mathcal{B}$ or all other types, respectively, and then the contrast, i.e. the gray values for black and white pixels. There is no further image preprocessing.

\section{System specifications}
The following part describes the general architecture of the entire workflow, and we explain details for the core neural network and the decoding procedure.

\subsection{Input}
\label{ss:Input}
The networks use entire writings as being prepared in the above preprocessing.
In particular, there is no further segmentation.
As developed in \cite{graves2008offline}, every writing image is processed by reading its pixel data in four column-first ``directions'' that arise from combining top-down and  bottom-up column traversals with left-to-right and right-to-left row traversals.

\subsection{Neural Network}
The neural networks used for CITlab systems are essentially based on preceding work presented in \cite{graves2008offline}.
The basics are the same for all network types applied for ANWRESH-2014.
However, as mentioned before, we use networks of various types and even several networks of the same type.
All those may be different in
\begin{itemize}
  \item the total (and per-layer) numbers of cells (neurons) they are built of, see \ref{sss:Architecture};
	\item their output layer and data as these correspond to the respective task, see \ref{sss:OutputType};
	\item the learning data and certain parameters used for train them, see \ref{ss:Training}.
\end{itemize}

\subsubsection{Architecture}
\label{sss:Architecture}
The architecture of the neural nets particularly follows \cite{graves2008offline}, but we introduced essential modifications:
Instead of traditional MDLSTM cells, we use two layers with MDLeaky cells (\cite{citlab2013openhart}) which were shown to be more stable and thus yield better performance.
They embrace one layer of classical $\tanh$ cells in between.

Furthermore, this network core is preceeded by a first layer accomplishing \textsc{Gabor} filtering with fixed parameters, thus essentially reducing the number of trainable weights and, consequently, training time.
Finally, the concluding layer right before the output layer simply adds up preceding activations such that for every pixel column, just one value arrives here. Thus, this results in an overall size reduction (subsampling) from the standard image height (see Section \ref{s:Preprocessing}) downto 1, but in fact, it is done step-wisely over the layers by factors 4 in y-direction or 3 in x-direction, respectively.

While the above details are common to all networks used here, for different networks, we tested various numbers of units. Those finally chosen are again based on \cite{graves2008offline}) and, particular, on \cite{citlab2013openhart} and further own experimental experience and testing.
Table \ref{tab:tableArch} shows some network architecture parameters in detail.
\begin{table}[ht]
  \centering
  \begin{tabular}[t]{ c @{\hspace{3em}} D{.}{}{0} *{2}{D{.}{}{3}} }
    & \multicolumn{3}{c}{number of} \\ \cline{2-4} \\[-.75em]
    network & \multicolumn{1}{c}{cells} & \multicolumn{1}{r}{trainable weights} & \multicolumn{1}{r}{output neurons} \\ \hline \\[-.75em]
    $\mathcal{N}_1$ & 1900 & 958387 & 56 \\
    $\mathcal{N}_2$ & 1166 & 363477 & 56 \\
    $\mathcal{N}_3$ & 1166 & 363477 & 56 \\
    $\mathcal{R}_1$ & 1900 & 1006387 & 56 \\
    $\mathcal{R}_2$ & 1900 & 1006387 & 56 \\
    $\mathcal{A}$ & 1869 & 933556 & 25 \\
    $\mathcal{M}$ & 1851 & 967138 & 7 \\
    $\mathcal{B}_1$ & 1899 & 1005586 & 55 \\
    $\mathcal{B}_2$ & 1532 & 655747 & 55 \\
  \end{tabular}
  \caption{Architecture parameters}
  \label{tab:tableArch}
\end{table}

\subsubsection{Output \& Type}
\label{sss:OutputType}
According to its specific tasks, every network type has a particular alphabet for the data to be read.
In the following table, we summarize the details. The shown number of output neurons also counts the standard, artificial ``garbage'' symbol that every alphabet additionally contains.

Note that one particularity in the \texttt{NAME} data field deserved special treatment: While in the images, family names were abbreviated by a solid line if they were just repeated for consecutive family members, the reference data contained the fully spelled family name. We reverted this by replacing the reference by an underscore \texttt{\_} as a special symbol.
\begin{table}[h]
	\centering
		\begin{tabular}[t]{c@{\hspace{3em}}cD{.}{}{3}l}
			network type & table data & \multicolumn{1}{r}{$\sharp$\,output neurons} & alphabet \\ \hline \\[-.75em]
			$\mathcal{N}$ & \texttt{NAME}& 56 & \verb*/ /'\_ A\dots Z a\dots z \\
			$\mathcal{R}$ & \texttt{RELATION} & 56 & \verb*/ /-\_ A\dots Z a\dots z \\
			$\mathcal{A}$ & \texttt{Age} & 25 & \verb*/ / 0\dots 9 $\frac{0}{12}$\dots $\frac{12}{12}$ \\
			$\mathcal{M}$ & \texttt{Marital condition} & 7 & letters S,M,W,D,C,V \\
			$\mathcal{B}$ & \texttt{PLACE OF BIRTH} & 55 & \verb*/ /- A\dots Z a\dots z \\
		\end{tabular}
	\caption{Network output}
	\label{tab:NetworkOutput}
\end{table}

\subsubsection{Training Algorithm} 
As usual, the activation of an output neuron at a particular time is trained to estimate the probability of the occurrence of its corresponding character at a specific position in the original writing.
The network is trained by Backpropagation-Through-Time (BPTT) using the Connectionist Temporal Classification (CTC) algorithm described in \cite{graves2006connectionist} to calculate the gradient.

\subsection{Decoding}
\label{ss:Decoding}
After applying the standard softmax normalization, at each time step, the neural net provides a vector of probability estimates, each component counting for one entry of the alphabet. Collecting those vectors over time, finally yields the network output matrix, ${N}(x)$, for a given input writing, $x$.
Decoding algorithms then typically search for a most likely word $w^*$ for the network output under consideration: 
\[w^*= \argmax_{w}p(w| {N}(x)) \,.\]
Since the garbage class typically has a high probability compared to other classes, the garbage ''letter'' is often cheap to insert which might mislead the decoding between shorter or longer word guesses. In order to correct for this, we prefer short words by ''normalizing'' the above word probability by some function depending on the word length. Furthermore, in order to incorporate prior probabilities, we multiply with relative frequencies, $p(w)$, computed from the ANWRESH dictionaries. All in all, we take  
\begin{equation}
\label{eq:MinCost}
w^* = \argmin_{w} \frac{-\ln p(w|{N}(x))}{|w|^\alpha} - \beta\ln p(w)
\end{equation}
where $\alpha$ and $\beta$ are some constants.

For some data fields, we apply several networks simultaneously, and choose the answer
\begin{equation}
\label{eq:MinCostMoE}
w^* = \argmin_{w,i\in I} \frac{-\ln p(w|{N_i}(x))}{|w|^\alpha} - \beta\ln p(w)
\end{equation}
where $I$ is the index set of all networks.
The value used in (\ref{eq:MinCostMoE}), $\min\limits_{i\in I}\frac{-\ln p(w|{N_i}(x))}{|w|^\alpha} - \beta\ln p(w)$,
(resp. (\ref{eq:MinCost}) if not applying committee's vote) 
will be called the \textsl{cost} of a certain dictionary entry $w$.

In a postprocessing stage, we delete inconsistent field results: If, e.g., the relation is \textit{wife} and the given name is doubtlessly male, we add additional penalty costs onto the costs of either the relation or the name result depending on which field is more unlikely. Let's say the name is more unlikely and gets additional penalty costs, then, if there is a sufficiently probable second  alternative being female, this second name will be preferred over the first name.

\section{Application}
We conclude with describing the working setup specifically used for this competition in training and testing our system.

\subsection{Training}
\label{ss:Training}
We divide the ANWRESH-2014 data set into a training and a validation set: For each data field type, these sets contain about 120k and 12k writings, respectively. The exact value depends on the image ground truth: Some of them are missing or for other reasons unusable such that some field images have to be omitted. In every epoch of the batch training, 20k of these writings were used. 

Networks were trained with a fixed momentum of 0.9, and two different learning rates: we started the main training portion with 0.002 and then concluded by a rather short post-training with a learning rate of 0.001.
Table \ref{tab:tableTrain} shows the various numbers of epochs that the used networks have been trained for. The intended maximum was 100 epochs but mainly due to time limitations, we eventually had to stop earlier.
\begin{table}[ht]
  \centering
  \begin{tabular}[t]{ c@{\hspace{3em}} *{2}{D{.}{}{3}} }
    & \multicolumn{2}{c}{number of training epochs} \\ \cline{2-3} \\[-.75em]
    network & \multicolumn{1}{r}{main learning} & \multicolumn{1}{r}{post-training} \\ \hline \\[-.75em]
    $\mathcal{N}_1$ & 31 & 8 \\
    $\mathcal{N}_2$ & 54 & 8 \\
    $\mathcal{N}_3$ & 60 & 9 \\
    $\mathcal{R}_1$ & 100 & 7 \\
    $\mathcal{R}_2$ & 100 & 10 \\
    $\mathcal{A}$ & 100 & 6 \\
    $\mathcal{M}$ & 100 & 9 \\
    $\mathcal{B}_1$ & 67 & 8 \\
    $\mathcal{B}_2$ & 99 & 8 \\
  \end{tabular}
  \caption{Training parameters}
  \label{tab:tableTrain}
\end{table}

\subsection{Testing}
For the experiments, we finally used 9 neural nets altogether.
For every network (resp. committee) reading one data field, we then have to choose the constants $\alpha, \beta$ for the cost calculation (\ref{eq:MinCost}, \ref{eq:MinCostMoE}), see \ref{ss:Decoding}.
Starting from values by experience, these parameters where roughly optimized by rather small grid searches over some validation data taken from the ANWRESH training data, see \ref{ss:Training}.

Here, it turned out that family names, i.e. the first word of the table field \texttt{NAME}, and given names, i.e. the rest of that field, should better use different decoding weights

Table \ref{tab:DecodingParameters} shows those committee and decoding parameters details.
\begin{table}[h]
	\centering
		\begin{tabular}[t]{cc@{\hspace{3em}}ccc}
			& & number of networks & decoding & weights \\ \cline{4-5}
			field data & network type & (size of committee) & $\alpha$ & $\beta$ \\ \hline \\[-.75em]
			\texttt{NAME} -- family name & \multirow{2}{*}{$\mathcal{N}$} & \multirow{2}{*}{3} & 0.50 & 0.50 \\
			\texttt{NAME} -- given name & & & 0.25 & 0.25 \\
			\texttt{RELATION} & $\mathcal{R}$ & 2 & 1.00 & 0.00 \\
			\texttt{Age} & $\mathcal{A}$ & 1 & 0.75 & 0.25 \\
			\texttt{Marital condition} & $\mathcal{M}$ & 1 & 0.75 & 0.25 \\
			\texttt{PLACE OF BIRTH} & $\mathcal{B}$ & 2 & 1.00 & 0.00 
		\end{tabular}
	\caption{Committees and Decoding Parameters}
	\label{tab:DecodingParameters}
\end{table}

%
%
%

\subsection*{Acknowledgement}
The work presented in this paper was funded by research grant no. V220-630-08-TFMV-S/F-059 (Verbundvorhaben, Technologieförderung Land Mecklenburg-Vorpommern) in European Social / Regional Development Funds.

\bibliographystyle{alpha}
\bibliography{SystemDescription_arXiv}

\end{document}